\pdfoutput=1
\documentclass[11pt]{article}

\usepackage[]{acl}

\usepackage{times}
\usepackage{latexsym}
\usepackage{graphicx} % Required for inserting images
\usepackage{amsmath}
\usepackage{amsfonts}
\usepackage{url}
\usepackage{booktabs}
\usepackage{subcaption}
\usepackage{multirow}
\usepackage{xcolor}

\usepackage[T1]{fontenc}
\usepackage[utf8]{inputenc}

\usepackage{microtype}

\usepackage{inconsolata}

\title{Mechanistic Understanding and Mitigation of Language Model Non-Factual Hallucinations}

\author{{Lei Yu$^{1, *}$, Meng Cao$^{2, 3, *}$, Jackie Chi Kit Cheung$^{2, 3}$, Yue Dong$^{4}$}\\
    $^1$Department of Computer Science, University of Toronto \\
    $^2$School of Computer Science, McGill University \\    
    $^3$Mila – Qu\'{e}bec AI Institute \\
    $^4$University of California, Riverside \\    
    {\tt jadeleiyu@cs.toronto.edu} \\
    {\tt  \{meng.cao@mail,jcheung@cs\}.mcgill.ca} \\
    {\tt  yue.dong@ucr.edu}
}

\begin{document}
\maketitle

\def\thefootnote{*}\footnotetext{Equal contribution.}
\renewcommand{\thefootnote}{\arabic{footnote}}

\begin{abstract}
State-of-the-art language models (LMs) sometimes generate \textit{non-factual hallucinations} that misalign with world knowledge. To explore the mechanistic causes of these hallucinations, we create diagnostic datasets with subject-relation queries and adapt interpretability methods to trace hallucinations through internal model representations. We discover two general and distinct mechanistic causes of hallucinations shared across LMs (Llama-2, Pythia, GPT-J): 1) \textbf{knowledge enrichment hallucinations}: insufficient subject attribute knowledge in lower layer MLPs, and 2) \textbf{answer extraction hallucinations}: failure to select the correct object attribute in upper layer attention heads. We also found these two internal mechanistic causes of hallucinations are reflected in external manifestations. Based on insights from our mechanistic analysis, we propose a novel hallucination mitigation method through targeted restoration of the LM's internal fact recall pipeline, demonstrating superior performance compared to baselines.

\end{abstract}
\section{Introduction}

Language models (LMs) serve as repositories of substantial knowledge \cite{petroni2019language,jiang2020can,srivastava2023beyond} through their parametric knowledge gained from pre-training. However, they are susceptible to generating ``hallucinations'' that contain factual errors. At the level of logit predictions, these hallucinations often display a pattern similar to factual generations. For example, LMs have been observed to produce seemingly confident completions that are, in reality, hallucinations \cite{dong2022survey,zhang2023siren}.

To understand how hallucinations differ from factual outputs and whether they are uniformly generated or equally challenging to fix, thorough analysis tools that monitoring the information flow are required, extending beyond merely last-layer predictions \cite{kaddour2023challenges}.  However, research on understanding the internal mechanisms of hallucination generation is limited. Most efforts on detecting and mitigating hallucinations \cite{elaraby2023halo, mundler2023self, manakul2023selfcheckgpt, zhang2023language} treat the LM as a black box,  devising methods based on external features like predictive uncertainty \cite{xiao-wang-2021-hallucination,varshney2023stitch} and logical consistency \cite{cohen2023lm}.  These approaches provide little insight into the internal mechanisms of factual errors and have been shown to be unreliable with often contradictory signals \cite{turpin2023language}.

In contrast, interpretability research, which examines the internal mechanisms of transformers in white-box settings, enables the identification of components that contribute to accurate factual predictions.  For example, existing work has identified several critical model ``components'' (e.g., attention heads, feedforward layers) related to knowledge flow that are essential for answering questions accurately \cite{lu2021influence,dai2022knowledge,meng2022locating,geva2023dissecting}.  However, it remains unclear whether the results of mechanistic interpretability on factual predictions can generalize to hallucinations. Specifically, it is \textit{unknown which model components deviate from normal functioning to cause hallucinations}. Localizing the source of non-factual hallucination in LMs may help us design targeted and efficient methods to mitigate hallucinations without significantly impacting utility (e.g., by editing a small set of model weights identified as causing hallucinations, without affecting other parts that are important for information flow). 

\begin{figure*}[t!]
\centering
\includegraphics[width=0.95\textwidth]{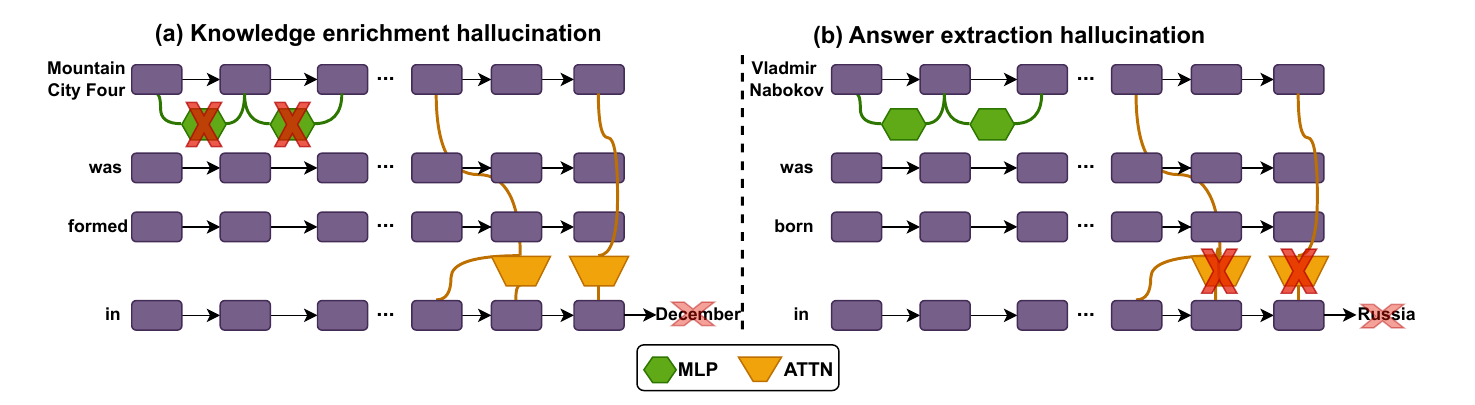}
\caption{\textbf{Our main finding of two non-factual hallucination mechanisms.} \textbf{Left (a)}: The \textbf{knowledge enrichment hallucinations} are caused by lacking general knowledge of the subject retrieved from early and middle layer MLPs -- in these cases, the subjects tend to be relatively unknown and the incorrect answer is often nonsensical. \textbf{Right (b)}: The \textbf{answer extraction hallucinations} are caused by the failure of middle and late layer self-attention heads to identify the most relevant object to the given subject and relation -- in these cases, the subjects are often more strongly associated with the hallucinating answers than the with the true answers.}
\label{fig:early-late-hall-illustrate}
\end{figure*}

In this study, we employ mechanistic interpretability \cite{olah2022mechinterp} to investigate the origins and manifestations of non-factual hallucinations in LMs. To address the lack of datasets for non-factual hallucinations, we constructed a diagnostic hallucination dataset from ParaRel \cite{elazar-etal-2021-measuring}, which contains cloze-style factual knowledge queries. This enables the examination of information flow associated with non-factual hallucinations.  Specifically, we adapt two established interpretability methods for hallucination analysis—logit lens \cite{geva-etal-2022-transformer,dar2023analyzing} and causal mediation analysis \cite{pearl2001direct,vig2020investigating}—aiming to assess the influence of model components on the generation of hallucinatory predictions. Through extensive analyses on LMs of various sizes and architectures (Llama-2, Pythia, GPT-J), we obtain converging evidence that there exist two groups of crucial components for factually incorrect predictions: 1) the multi-layer perceptrons (MLPs) in lower transformer layers, 2) the multi-head self-attentions in upper transformer layers. 

Figure~\ref{fig:early-late-hall-illustrate} illustrates two distinct scenarios where the identified hallucinating components exhibit different behaviors. In some instances, lower-layer MLPs function normally, successfully retrieving semantic attributes about queried entities, while upper-layer attention heads struggle to distinguish the most relevant attributes that lead to the correct answer. In other cases, the model fails to execute its fact-recalling pipeline at the beginning, extracting no useful information from lower-layer MLPs. We also observe that these two hallucination mechanisms have varying external manifestations, distinguishable by their levels of subject-object association strengths, robustness to input perturbations, and model predictive uncertainty. Moreover, we demonstrate that the \textbf{mechanistic insights gained from our analyses can be leveraged to develop an effective method to reduce LM hallucinations} on multiple open-domain question answering datasets. Our research offers the first mechanistic explanation and mitigation of LM factual errors, fostering future research on model explainability and transparency.

% Following the discoveries by \citet{akyurek2022towards,zhou2023lima} regarding the significant role of LM pre-training in acquiring factual knowledge, we also examined the impact of pre-training on the various types of hallucinations we have identified. Our findings are significant: early-site MLPs and late-site attentions form a two-step pipeline for fact recall that emerges progressively during pre-training, and failing to develop either step can result in hallucinations. 
% Moreover, as a practical application, we demonstrate that mechanistic interpretability features can be employed to probe the presence of factual errors in LMs. Our work offers the first mechanistic explanation of LM factual errors as systematic modular failures, fostering research on model explanability and hallucination mitigation. 
% \yue{I guess we have discussed this multiple times, if external features can predict, why we need to see internal ones?}

% Change this paragraph to hallucination mitigation
% In addition to providing internal knowledge flow analysis that exhibits high correlations with external features, which helps us understand how different types of hallucinations behave during various stages of pre-training, our work also demonstrates a practical application: mechanistic interpretability features can be employed to detect the presence of factual errors in language models (LMs). This research offers the first mechanistic explanation of LM factual errors as systematic modular failures, fostering research on model explainability and hallucination mitigation.

\section{Related Work and Background}   \label{sec:background}
\paragraph{Factual knowledge in language models.}
The exploration of knowledge tracing within Language Models (LMs) has gained substantial attention lately, with researchers investigating specific layers \citep{singh-etal-2020-bertnesia,geva2021transformer,meng2022locating} and neurons \citep{dai2022knowledge} responsible for storing factual information. This line of inquiry extends to techniques for model editing \citep{de2021editing,mitchell2021fast,meng2022mass} and inference intervention \citep{hernandez2023linearity,li2023inference}. Recent advancements by \citet{geva2023dissecting,yu2023characterizing} identify crucial LM components that form an internal pipeline for factual information transfer. Our framework complements existing research by offering an additional perspective on LM factual knowledge processing, revealing that compromised factually relevant modules can lead to hallucinations.

\paragraph{Hallucinations.}
Language models are susceptible to generating hallucinations that can be \textit{unfaithful} (i.e. deviating from the source input provided by users) or \textit{non-factual} (i.e. contradicting established world knowledge) \cite{cao-etal-2022-hallucinated,ji2023survey,zhang2023siren}. Here, we focus on the latter type of hallucination. Existing studies aimed at detecting or mitigating hallucinations leverage features such as internal activation patterns \cite{yuksekgonul2023attention,li2023inference}, predictive confidence \cite{cao-etal-2022-hallucinated, cao-etal-2022-learning, varshney2023stitch}, and generation consistency \cite{mundler2023self,manakul2023selfcheckgpt,zhang2023language}.
% However, a mechanistic investigation accounting for non-factual hallucinations is lacking in these studies. 

\paragraph{Mechanistic interpretability.}
Mechanistic interpretability \cite{olah2022mechinterp,nanda2023attribution} is an evolving research area. Recent works employ projections to the vocabulary \cite{geva-etal-2022-transformer,geva2022lm,nostal2020logitlen,katz2024backward} and interventions in transformer computation \cite{finlayson2021causal,haviv-etal-2022-transformer,stolfo2023mechanistic,ghandeharioun2024patchscope} to study LM inner workings,  explore neural network learning dynamics \cite{nanda2022progress} and discover sparse computational graphs for specific tasks \cite{wang2022interpretability,conmy2023towards}. Leveraging multiple mechanistic interpretability methods, our study provides a principled account and mitigation method for non-factual hallucinations.

% \section{Background and Notation} \label{sec:background}

\paragraph{Background and notations} Our work builds on the inference pass of decoder-only, transformer-based LMs. An auto-regressive transformer~\citep{vaswani2017attention}, denoted as $G$, maps an input sequence of tokens $u = [w_1,...,w_T]$ into a probability distribution over the vocabulary for next-token prediction. Within the transformer, the $i$-th token is represented as a series of hidden states $h_i^{(l)}$ where at each layer $l$, the model computes and adds the intermediate embeddings by two modules from $h_i^{(l-1)}$: 1) an aggregated \textbf{multi-head self-attention module} output $a^{(l)}_i = W_o([a_{i}^{(l,0)},...,a_{i}^{(l,K)}])$, where $a_{i}^{(l,k)}$ is the output of the $k$-th attention head at layer $l$ (with $K$ heads in total) for the $i$-th token \footnote{$a_{i}^{(l,k)}=\text{softmax}\left( \frac{Q^k_i(K^k_i)^T}{\sqrt {d} }\right) \cdot V^k_i$ and $Q^k_i,K^k_i,V^k_i$ are derived from $h_i^{(l-1)}$ with linear transformations.}, and $W_o$ is a linear transformation; 2) a \textbf{multi-layer perceptron (MLP)} output $m_i^{(l)}$ = $f_\text{MLP}^{(l)}(h_i^{(l-1)} + a^{(l)}_i)$ at layer $l$. Putting together, the hidden representation $h_i^{(l)}$ is computed as: $h_i^{(l)} = h_i^{(l-1)} + a^{(l)}_i + m_i^{(l)}$. Let $H = \{h^{l}_{i}\}$ be the set of $T \times L$ token hidden states across all layers (following \citet{elhage2021mathematical}, we shall call them the \textbf{residual stream outputs}), $A = \{a^{l}_{i}\}$ be the set of $T \times L$ \textbf{attention outputs}, and $M = \{m^{(l)}_i\}$ be the set of $T \times L$ \textbf{MLP outputs}. We aim to investigate which intermediate hidden representations $z \in Z = A \bigcup M$ are causing the model to generate a factually incorrect answer for an input question. %For further details on transformers, see \citet{vaswani2017attention}. 

\section{Dataset for LM Hallucination} \label{sec:hall-datasets}

% \section{Dataset \& Mechanistic Analysis of Hallucinations} \label{sec:hall-mech-analysis}

% \begin{table}[]
% \centering
% \resizebox{0.85\columnwidth}{!}{%
% \begin{tabular}{@{}llll@{}}
% \toprule
%  & Llama & Pythia & GPT-J \\ \midrule
% Factual set size & 25204 & 10277 & 8646 \\
% Hallucinating set size & 25478 & 31110 & 23831 \\
% Model accuracy & 0.497 & 0.248 & 0.266 \\ \bottomrule
% \end{tabular}%
% }
% \caption{Statistics of the ParaRel query datasets of three language models.}
% \label{tab:pararel-ds-stats}
% \end{table}
% \usepackage{booktabs}
% \usepackage{graphicx}

\paragraph{Dataset Construction}
We collect a set of questions from the ParaRel \cite{elazar-etal-2021-measuring} dataset of cloze-style factual knowledge queries. Each example in ParaRel consists of a subject-relation-object triple $(s, r, o)$ (e.g., (\textit{Paris}, CAPITAL\_CITY, \textit{France})) and a set of prompts $u(s,r,o)$ generated from hand-curated templates that contains $(s,r)$ and has $o$ as its ground-truth next word continuation (e.g., ``The capital city of France is ''). To ensure the uniqueness of the true answer for each query, we only take prompts generated from triples in the ``many-to-one'' relational classes in ParaRel where each subject-relation has a single associated object entity that begins with a capitalized English letter. This yields a large set of approximately 80K factual knowledge queries. 

We evaluated three widely used pretrained LMs on our constructed query dataset: 1) Llama-2 (32 layers, 7B parameters, fine-tuned on instruction following) \cite{touvron2023llama}, 2) Pythia \cite{biderman2023pythia} (32 layers, 6.9B parameters), and 3) GPT-J (28 layers, 6B parameters) \cite{gpt-j}. For each prompt $u$, we compute the LM predicted conditional probability $p(t|u)$ of the next token continuation, where $t$ is taken from the collection of all capitalized alphabetical tokens in the model vocabulary. We define the \textbf{non-factual hallucination set} as the queries for which a model predicted next token $\hat{t} = \underset{t}{\text{argmax }} p(t|u)$ is not a prefix of the true object answer $o$ (i.e., the model makes a factual error), and otherwise the query is an example of the \textbf{factual set} (i.e., the model answers correctly). Finally, for each model, we discard those queries with no capitalized alphanumeric tokens among model predicted top-50 most likely tokens over the entire vocabulary, as we found in most of these cases the log likelihood of $\hat{t}$ would become negligible and therefore not suitable for our subsequent analyses. Table \ref{tab:pararel-ds-stats} summarizes the dataset statistics.

\section{Mechanistic Analysis of Hallucinations} \label{sec:hall-mech-analysis}

\begin{table}[]
\centering
\resizebox{.95\columnwidth}{!}{%
\begin{tabular}{@{}l|lll@{}}
\toprule
 & \textbf{Llama-2} & \textbf{Pythia} & \textbf{GPT-J} \\ \midrule
No. of factual queries & 25204 & 10277 & 8646 \\
No. of hallucinating queries & 25478 & 31110 & 23831 \\
Model accuracy & 0.497 & 0.248 & 0.266 \\ \midrule
\% of enrichment hall. & 22.1 & 30.2 & 67.3 \\
\% of extraction hall. & 77.9 & 69.8 & 32.7 \\ \bottomrule
\end{tabular}%
}
\caption{Statistics of the ParaRel query datasets of three language models.}
\label{tab:pararel-ds-stats}
\end{table}
We wish to know which ``broken'' LM components are causing the model to produce a factually incorrect answer. Recent studies have shown that given a subject-relation query, an LM predicts a factual object answer via two steps \citep{geva2023dissecting,yu2023characterizing,jin2024cutting}: 1) during the \textbf{knowledge enrichment step}, the model retrieves from MLP sublayers many relevant semantic attributes of the subject and propagates them to the last query token position; and 2) during the \textbf{answer extraction step}, the self-attention modules select the most relevant object entity among the previously retrieved attributes. We postulates that an LM hallucinates if any one of these two steps get compromised during inference, and perform a series of mechanistic interpretability analyses to pinpoint the malfunctioning model components. 

\subsection{Model inspection through logit lens}
\paragraph{Method}
We inspect the semantic information encoded in the intermediate hidden representations within each transformer layer through logit lens \citep{nostal2020logitlen,elhage2021mathematical,dar2023analyzing}. In particular, for each $z^{(l)}_i \in A \bigcup M$ produced by either the MLP or the self-attention module at layer $l$ when processing the $i$-th query token, we cast it into a probability distribution over the LM vocabulary space by passing $z^{(l)}_i$ directly through the last prediction head layer: 
\begin{align}
    p(z_i^{(l)}) =  \text{softmax}(E \text{ LayerNorm}(z^{(l)}_i) ); 
\end{align}
where $E \in \mathbb{R}^{|V|\times d}$ is the unembedding matrix, and $\text{LayerNorm}$ is the layer norm operation. 

To quantify the information of the true answer $o$ that an LM extracts when processing the subject tokens at each layer, we compute the logit values  $\mathcal{I}_m^{(l)}(o) = e_o^T \text{ LayerNorm}(m^{(l)}_s)$ of projecting the MLP-produced hidden representation of the last subject token $m^{(l)}_s$ onto the unembedding vector $e_o$ of the object token. A high value $\mathcal{I}_m^{(l)}(o)$ would indicate that the model has already enriched the subject with sufficient information of an object before processing the relation. Similarly, given the true object token $o$ and another set of attribute tokens $o' \in O'$ with high MLP-enriched information $\mathcal{I}_m^{(l)}(o')$ when processing $s$ \footnote{We take the top-100 tokens in model vocabulary with highest $\mathcal{I}_m^{(l)}(o')$ as $O'$.}, we can measure how good the self-attention module in layer $l$ is at distinguishing $o$ against other attributes $o'$ by computing the relative attention-extracted attribute information $\mathcal{I}_a^{(l)}(o) = a^{(l)}_T \Big(e_o  -  \Bar{e}_{o'}\Big)$, where $a^{(l)}_T$ is the attention module output when processing the last input token, and $\Bar{e}_{o'} = \frac{1}{|O'|}e_{o'}$ is the mean unembedding vector of the non-answer attributes. A high value of $\mathcal{I}_a^{(l)}(o)$ suggests that the attention modules can effectively identify $o$ as the target attribute when synthesizing information propagated from subject and relation tokens. 

\begin{figure}[]
\centering
\includegraphics[width=0.9\columnwidth]{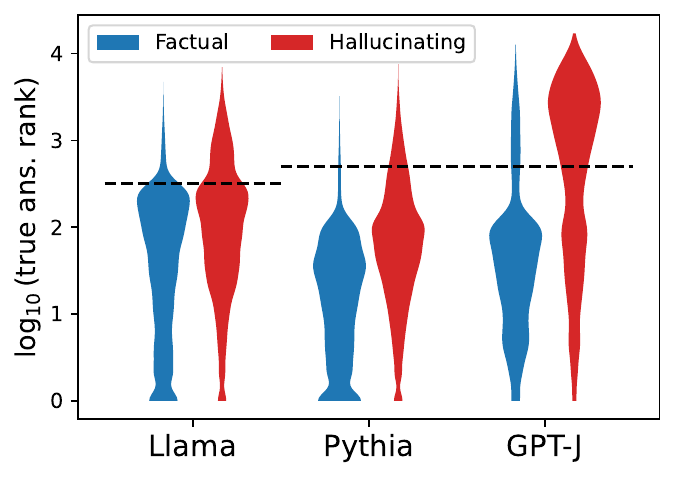}
\caption{Minimum (over all transformer layers) true object token rankings in the logit lens distributions of intermediate MLP outputs (shown in log scale). Dashed lines denote the threshold $\rho^*_s(o) =0.01|V|$ ranks to distinguish between knowledge enrichment and answer extraction hallucinations ($\rho^*_s(o)=320$ for Llama-2 and $\rho^*_s(o)=502$ for Pythia/GPT-J).}
\label{fig:pararel_ans_rankings}
\end{figure}

\begin{figure*}[ht]
\centering
\includegraphics[width=0.9\textwidth]{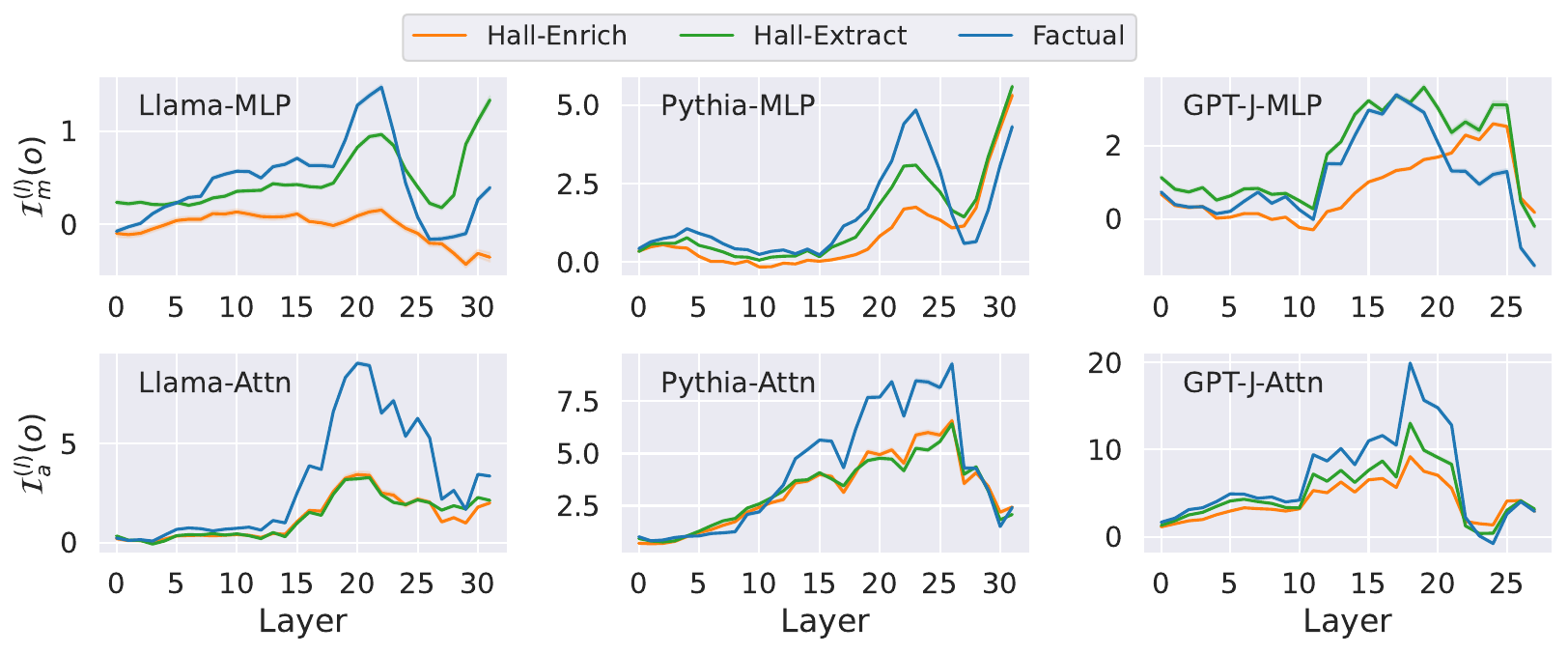}
% \caption{\textbf{Average Indirect Effect (AIE)} of individual model components to non-factual hallucinations over 6,401 ParaRel queries that are incorrectly answered by GPT-2 XL. $\Delta \text{AIE}(y,u)$ is defined as the difference in AIE between 1) the attention outputs of the last 24 transformer layers and 2) the MLP outputs of the first 24 GPT-2 XL layers.}
\caption{Average logit lens projection values between true object embedding and intermediate MLP/attention representations of Llama-2/Pythia/GPT-J in each transformer layer.
}
\label{fig:pararel_logit_lens}
\end{figure*}

\paragraph{Two mechanisms of hallucinations}
We first examine whether the LMs retrieve sufficient information about the answer during the subject knowledge enrichment process. We consider an attribute to be sufficiently extracted from the model parametric knowledge base if it is among the top 1\% tokens of highest MLP-retrieved information $\mathcal{I}_m^{(l)}(\cdot)$ in at least one some intermediate layers. For each query $u(s,r,o)$ in the factual and the hallucinating set, we compute the minimum ranking $\rho^*_s(o)$ of $o$ in the logit lens distribution $p(m_s^{(l)})$ of MLP outputs across all LM layers, as shown in Figure \ref{fig:pararel_ans_rankings}. We observe that for the vast majority of factual set queries, there is at least one intermediate MLP representation in which the object ranks among top 1\% of the entire vocabulary. In contrast, for a significant portion of hallucinating examples, even the object token with the most MLP-retrieved information remains outside the top 1\% of the vocabulary.

We hypothesize that, for queries with $\rho^*_s(o) > 0.01|V|$, the model hallucinations are mostly caused by the insufficient knowledge extracted from MLPs, and we therefore call these examples \textbf{(knowledge) enrichment hallucinations}. On the other hand, for queries with $\rho^*_s(o) \leq 0.01|V|$, the model functions normally at the knowledge enrichment step, but later on fails at the answer extraction step where it cannot distinguish the object entity against the other related attributes of the subject, so we call these examples \textbf{(answer) extraction hallucinations}. The last two rows of Table \ref{tab:pararel-ds-stats} shows the percentage of queries that fall into each hallucination type for the three models. We noticed that the majority of Llama-2 and Pythia errors are extraction hallucinations, while GPT-J have much more enrichment hallucinations, suggesting that GPT-J may suffer from a more severe lack of general world knowledge compared to more recent LMs.  

To better understand the two identified hallucination mechanisms, we compute the average layerwise MLP-enriched and attention-extracted object information for factual queries and hallucinating queries with the two error types, as illustrated in Figure \ref{fig:pararel_logit_lens}. Some key observations are: 1) both factual queries and extraction hallucinations retrieve a significant amount of object information from MLPs in early and middle transformer layers, whereas  enrichment hallucinations have much less object knowledge incorporated into the subject tokens in early inference stages. 2) Compared to the factual query set, the self-attention module outputs of both types of hallucinations fail to effectively distinguish $o$ against other incorrect attributes. 

These findings together suggest that failures of either MLP knowledge enrichment or self-attention answer extraction would cause non-factual hallucinations. Moreover, sufficient retrieval of object knowledge in early layers serves as a prerequisite of answer extraction, so a degenerated enrichment process will inevitably compromise the ability of upper-layer attention to filter irrelevant attributes, as observed in enrichment hallucinations.

\begin{figure*}[ht]
\centering
\includegraphics[width=.95\textwidth]{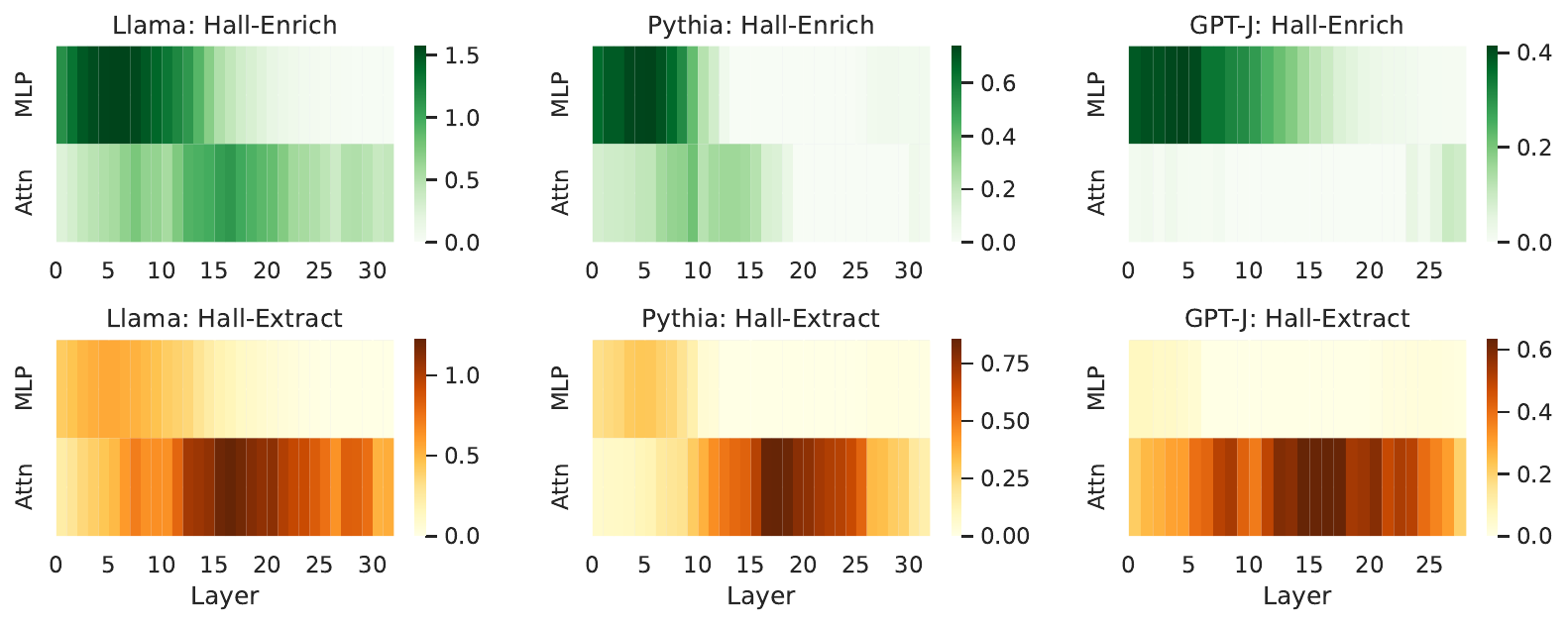}
% \caption{\textbf{Average Indirect Effect (AIE)} of individual model components to non-factual hallucinations over 6,401 ParaRel queries that are incorrectly answered by GPT-2 XL. $\Delta \text{AIE}(y,u)$ is defined as the difference in AIE between 1) the attention outputs of the last 24 transformer layers and 2) the MLP outputs of the first 24 GPT-2 XL layers.}
\caption{\textbf{Average Indirect Effect (AIE)} of mitigating MLP and self-attention intermediate outputs for (a) enrichment hallucinations (green heatmaps) and (b) extraction (orange heatmaps) hallucinations. 
}
\label{fig:pararel_causal_IEs}
\end{figure*}

\subsection{Causal validation of hallucination mechanisms}
\label{subsec:causal_patching}
If lower layer MLP and upper layer self-attention outputs are the root causes of non-factual hallucinations, then fixing them should enhance model factuality. We test this hypothesis by performing a causal patching analysis to measure the contribution of each intermediate representation to a hallucinating model prediction. 

\paragraph{Method}
The intermediate hidden states produced by an LM during inference form a causal dependency graph \cite{pearl2001direct} that contains many paths from the input sequence to the output (next-token prediction), and we wish to understand if there are specific hidden states that are more important than others when producing a hallucination. This is a natural case for \textit{causal mediation analysis} \cite{vig2020investigating,meng2022locating}, which quantifies the contribution of intermediate variables in causal graphs. Given a query $u(s,r,o)$ and a model generated incorrect object $o'$, we consider the LM as a ``corrupted'' model with certain modules failing to compute the ``clean'' representations that could otherwise lead to the correct answer $o$, and measure the contribution of each module through three model runs: 
\begin{enumerate}
    \item In the first run, we pass $u$ into the model and extract all intermediate hidden representations $z$ as defined in Section~\ref{sec:background}, and compute the log likelihood ratio $y = \log\frac{p(o'|u)}{p(o|u)}$ between the true and hallucinated objects, which quantifies the model's ``degree of hallucination'' when answering $u$. For a hallucinating prediction, we would observe $y > 0$.  
    \item In the second run, we inject a Gaussian noise $\epsilon \sim \mathcal{N}(0, \sigma)$ \footnote{We follow \cite{meng2022locating} to set $\sigma$ to be be 3 times lof the empirical standard deviation of the input embeddings.} to the subject token embeddings of $u$. Let $u^*$ denote the resulting query with perturbed input embeddings, we re-compute the log-likelihood ratio $y' = \log\frac{p(o'|u^*)}{p(o|u^*)}$ and take those noises with $y' < 1$ (i.e., we only keep noises that make the model become truthful by preferring $o$ over $o'$). We again extract all intermediate hidden representations $z^*$. 
    \item In the third run, we again provide the model with $u^*$ with perturbed input embeddings, and ``patch'' a particular hidden representation $z^*$ to be the hidden representation $z$ during the first run. We then compute the log likelihood ratio $y'' = \log\frac{p(o'|u^*,z)}{p(o|u^*,z)}$ to see how it changes compared to step 2. 
\end{enumerate} 
If an intermediate output $z$ is the main cause of a hallucination, then overwriting it with $z^*$ produced during a truthful run should also make the model more factual. We define the causal \textbf{indirect effect} $\text{IE}(z; y, u, \epsilon) = y'' - y$ of $z$ as the decrease in the degree of hallucination after mitigating a single hidden state. Averaging over a set of factual queries and a sample of noises for each query, we obtain the average indirect effect (AIE) for each $z$ and its corresponding MLP or self-attention component. 

\paragraph{Causal tracing results}
We compute layerwise AIE for MLP intermediate outputs $m_s^{(l)}$ of the last subject token and the self-attention intermediate outputs $a^{(l)}_T$ of the last input token for each hallucinating query. Figure \ref{fig:pararel_causal_IEs} shows the average MLP and attention causal effects to enrichment and  extraction hallucinations respectively. We observe a clear distinction between the causal contribution distributions of the two hallucination mechanisms: in particular, most intermediate hidden states that contribute significantly to enrichment hallucinations are produced by lower layers MLPs when processing subject tokens, and extraction hallucinations are mostly caused by outputs of upper layer self-attention heads right before generating the answer tokens. These findings conform with our logit lens analyses results, and together suggest that 1) \textit{lower layer MLPs and upper layer self-attention heads are the ``brittle'' LM components which, if compromised, would lead to non-factual hallucinations}, and 2) \textit{lower layer MLPs and upper layer attentions do not always break down simultaneously, thus leading to two distinct mechanisms of LM factual errors.}

\begin{table}[]
\centering
\resizebox{0.95\columnwidth}{!}{%
\begin{tabular}{@{}cccc@{}}
\toprule
\multicolumn{1}{c}{Statistics} & Know.Enrich. Hall. & Ans.Extract. Hall. \\ \midrule
$s$-$o$ assoc.                            &    0.47         &     1.17       \\
$s$-$o'$ assoc.                           &    0.81         &    1.69        \\
Robustness                 &     0.86        &    0.45        \\
Uncertainty                   &     4.54        &    4.76        \\ \bottomrule
\end{tabular}%
}
\caption{External data and model prediction features for two types of non-factual hallucination, averaged over three LMs.}
\label{tab:hall-mech-manifest-llama}
\end{table}

\subsection{External manifestations of hallucination mechanisms.} 
To ensure our categorization of hallucinations isn't just a fabricated dichotomy based on internal computation patterns, we also explore external features to distinguish between the two types. We consider the following features: the \textbf{subject-object association strength} is measured as the inner product between the input layer embeddings of a subject $s$ and a true object $o$ or a hallucinating object $o'$; the \textbf{robustness} of a predicted object $o'$ is measured as the percentage of Gaussian noise injected during the mitigation run in section \ref{subsec:causal_patching} which, after being added to the input embeddings, fails to make the model prefer the true answer $o$ over $o'$; the \textbf{uncertainty} of model prediction is measured by the entropy of conditional next-token distribution $p(o|u)$. Table \ref{tab:hall-mech-manifest-llama} summarizes the results with external measures averaged over the three tested LMs. 

We found that 1) subjects of extraction hallucinations often have hallucinating objects of much stronger association strengths than true objects, so that the model fail to ``offset'' the prior propensity of model predicting $o'$ upon seeing $s$. Subjects of enrichment hallucinations, on the other hand, often have much weaker associations with both true and hallucinating objects; 2) extraction hallucinations are significantly less robust under input perturbations, probably because the model has already retrieved the correct object from early layers and is just ``one step away'' from distinguishing it against less relevant attributes; 3) the model is less certain about its predictions when generating enrichment hallucinations, a pattern that is consistent with previous findings that epistemic hallucinations (i.e., hallucinations due to lack of general world knowledge) are often associated with high predictive uncertainty \cite{xiao-wang-2021-hallucination}.

\section{Mechanistic Hallucination Mitigation} \label{sec:fact_error_correct}
In this section, we propose a novel Mechanistic Hallucination Mitigation (MHM) method that draws inspiration from our mechanistic analysis, and demonstrate that it can improve LM factuality in open-domain question answering.

\paragraph{Method} 
Given a question $x$ and its true answer $y$ (we take the first token for answers with multiple tokens), we wish to fix the model's imperfect knowledge enrichment and answer extraction modules in the fact-recalling pipeline when it generates an incorrect answer $y'$. We do so by encourage the LM to retrieve more information of $y$ from MLPs, and to suppress the information propagation of $t'$ from self-attention heads. In particular, during model inference with input $x$, we take the intermediate MLP outputs $m_i^{(l)}$ and attention head outputs $a_i^{(l)}$ within a specific layer range $l\in L_{m}$ or $l\in L_{a}$, and then project them directly to the language modeling head layer. Let $\log p_m^{(i)}(y|x), \log p_a^{(i)}(y|x)$ be the resulting log-likelihoods of $y$ in the projected distribution, we fine-tune the LM to optimize the following objective function:

\begin{small}
\begin{equation}
\mathcal{L}_\text{MHM}(x,y,y') = -\sum_{l\in L_{m}} \log p_m^{(l)}(y|x) - \sum_{l\in L_{a}} \log \frac{p_a^{(l)}(y|x)}{p_a^{(l)}(y'|x)}
\label{eq:MHM-objective}
\end{equation}
\end{small}
In practice, we find that $\mathcal{L}_\text{MHM}$ can be combined with the regular negative log likelihood (NLL) loss of LM fine-tuning to achieve the best factuality:

\begin{small}
\begin{equation}
    \mathcal{L}(x,y,y') = \mathcal{L}_\text{NLL}([x;y]) + \lambda \mathcal{L}_\text{MHM}(x,y,y')
\label{eq:MHM-objective-with-SFT}
\end{equation}
\end{small}
where $\mathcal{L}_\text{NLL}([x;y])$ is the average NLL loss of the concatenated sequence of a question and its true answer, and $\lambda$ is a hyperparameter that controls the relative importance of two loss terms.

\paragraph{Data and models.} 
We study non-factual hallucination mitigation of the three LMs we analyzed on two open-domain question answering benchmarks: 1) the \textbf{Natural Questions} dataset \cite{kwiatkowski2019natural} that consists of Google search engine queries annotated with answers and supporting Wikipedia pages, and 2) the \textbf{TruthfulQA} dataset by \citet{lin2022truthfulqa} consisting of adversarially constructed commonsense reasoning questions to measure whether an LM is truthful in generating answers. For Natural Questions, we asked each LM to generate up to 20 tokens conditioned on each question, and label the model generation as correct if it contains an exact match of the true answer. For TruthfulQA, where each question is paired with a set of ``plausible sounding but false'' answers, we evaluate each LM under a multiple-choice scheme by computing the average conditional likelihood per token for each candidate answer, and define a correct prediction as the case where an LM assigns highest average likelihood for the true answer. 
Following the evaluation scheme in LM knowledge editing and factual error rectification, we would expect a mitigation method to significantly reduce model hallucination without compromising its originally possessed knowledge. We therefore take examples of both benchmarks on which an LM produces an incorrect answer as our training set, and then construct two evaluation datasets: the \textbf{effectiveness evaluation set} consists of the GPT-4-generated paraphrases of each training question on which the original model hallucinates, and the \textbf{specificity evaluation set} are the original benchmark questions that the LMs correctly answers. A good mitigation method should therefore achieve high accuracy on both evaluation sets. 

\paragraph{Baseline Methods.} 
We evaluate MHM against several baseline methods that have shown promising results in model editing or factualilty improvement: 1) the vanilla supervised fine-tuning method without the MHM objective, 2) a 5-shot in-context learning (ICL) method of prompting the model with five examplar (question, true answer) pairs, 3) the MEND algorithm for knowledge editing \citep{mitchell2021fast} that learns a hypernetwork to perform targeted weight updates on knowledge-intensive LM parameters, and 4) DoLa \cite{chuang2023dola}-a decoding algorithm by contrasting the differences in logits obtained from projecting the later transformer layers versus earlier layers to the vocabulary space. We use default hyperparameters and experimental setups taken from their official implementations, and report the evaluation results on the same test datasets of MHM. \footnote{See Appendix \ref{appendix:hall-mitigate} for additional details. }

\begin{table}[t]
% \small
\centering
\resizebox{\columnwidth}{!}{%
\begin{tabular}{l|cc}
\toprule
             & \textbf{Natural Questions}     & \textbf{Truthful QA}           \\
             & Eff./Spec. (\%) & Eff./Spec. (\%) \\ \midrule
\multicolumn{3}{c}{\textbf{Llama-2-7B-chat}   }          \\
\midrule
ICL (5-shot) & 27.5 / 91.7                      & 36.8 / \textbf{97.9 }                     \\
SFT          & 41.8 / 82.3                      & 44.1 / 96.5                      \\
MEND         & 39.8 / 86.9                      & 33.7 / 40.5                      \\
DoLa         & 27.0 / 69.6                      & 43.4 / 81.0                      \\
MHM (Ours)   & \textbf{47.6 / 95.5}             & \textbf{48.2} / 95.6             \\ \midrule
        \multicolumn{3}{c}{\textbf{Pythia-6.9B}   }   \\ \midrule 

ICL (5-shot) & 22.6 / \textbf{98.5}                      & 28.3 / 92.0                      \\
SFT          & 37.9 / 89.9                      & 34.7 / \textbf{96.7}                      \\
MEND         & 34.3 / 91.8                      & 25.4 / 50.4                      \\
DoLa         & 23.8 / 66.5                      & \textbf{46.6} / 88.4             \\
MHM (Ours)   & \textbf{46.9} / 92.4             & 45.9 / 93.5                      \\ 
\midrule
\multicolumn{3}{c}{\textbf{GPT-J}   } \\
\midrule
ICL (5-shot) & 13.8 / 74.9                      & 30.6 / \textbf{96.1}                      \\
SFT          & 34.4 / 86.0                      & 46.0 / 92.3                      \\
MEND         & 36.7 / 89.2                      & 33.3 / 83.7                      \\
DoLa         & 16.8 / 71.7                      & 37.9 / 90.0                      \\
MHM (Ours)   & \textbf{43.8 / 89.4}             & \textbf{49.7} / 95.8    \\  
\bottomrule
\end{tabular}%
}
\caption{Evaluation results of various hallucination mitigation methods on Natural Question and Truthful QA. \textbf{Effectiveness} refers to model accuracy on paraphrased training questions, and \textbf{specificity} denotes the percentage of left-out questions on which the LM still remains truthful after applying a mitigation method. }
\label{tab:hall-mitigate-results}
\end{table}

\paragraph{Results.}
Table \ref{tab:hall-mitigate-results} shows model accuracy on paraphrase and specificity evaluation sets. We found that in all setups, MHM either yields the most effective mitigation results, or achieves a performance that is comparable to the best mitigation method. Meanwhile, MHM in most cases preserves more than 90\% of model performance on the specificity evaluation sets, indicating that our mechanistic mitigation of hallucinations does not compromise LMs' general world knowledge. In contrast, other baselines often yield inferior performance on at least one of the two datasets. In particular, knowledge editing methods such as MEND struggles at Truthful QA on which an LM often hallucinates due to failing to distinguish between the true answer and confusing distractors, while decoding rectification methods such as DoLa help little on model errors on Natural Questions that are often caused by insufficient knowledge about query entities. These results suggest that extensive reparation of the entire LM fact-recalling pipeline is essential for effective and specific mitigation of non-factual hallucinations.

\section{Conclusion}
We conducted various interpretability analyses on non-factual hallucinations made by language models. We show that both lower layer MLPs and upper layer attention heads in the model factual knowledge recalling pipeline may operate abnormally during model inference, thereby leading to two distinct mechanisms of LM factual errors: insufficient knowledge enrichment and ineffective answer extraction. Leveraging these insights, we proposed an effective method of LM hallucination mitigation. Our work establishes a mechanistic understanding of LM factual errors, and may inspire future research on explainable approaches of improving the reliability of language models. 

\newpage
% \clearpage
\section{Limitation}
Our study bears several limitations. Firstly, certain experiments depend on interpreting intermediate layer representations and parameters through projection to the vocabulary space via logit lens. While widely used, this method only approximates the encoded information of model components, particularly in early layers. Future work should consider more sophisticated methods such as Tuned Lens \cite{belrose2023eliciting} to probe information encoded in LM layers. Secondly, our focus on analyzing non-factual hallucinations with simple input sequences may not fully capture real-world LM behavior. Future investigations should apply mechanistic interpretability methods to study more complex and naturalistic contexts, considering longer input queries and potential adversarial features that may distract LMs from their normal inference processes.

\bibliography{custom}

\clearpage

\appendix
\appendix

\section{Full list of ParaRel relational classes}
See Table \ref{tab:pararel-relations} for a complete list of N-to-1 ParaRel relational classes and sample queries that we used to construct our mechanistic hallucination analysis dataset.

\begin{table*}[]
\caption{PARAREL relations with unique object answers and sample queries.}
\centering
\resizebox{\textwidth}{!}{%
\begin{tabular}{@{}lllll@{}}
\toprule
\multicolumn{1}{c}{Relation ID} & \multicolumn{1}{c}{Relation} & \multicolumn{1}{c}{No. of queries} & \multicolumn{1}{c}{Sample Query}                    & \multicolumn{1}{c}{True answer} \\ \midrule
P103                            & native language              & 977                                & The mother tongue of Victor Horta is                & Dutch                           \\
P138                            & named after                  & 645                                & Rawlings Gold Glove Award, which is named for       & glove                           \\
P159                            & headquarters location        & 967                                & The headquarter of Strait Shipping is located in    & Wellington                      \\
P176                            & manufacturer                 & 982                                & Honda RA272 is produced by                          & Honda                           \\
P264                            & record label                 & 429                                & Johnny Carroll's record label is                    & Decca                           \\
P279                            & subclass of                  & 964                                & Nucleoporin 62, a type of                           & protein                         \\
P30                             & continent                    & 975                                & Romulus Glacier is located in                       & Antarctica                      \\
P407                            & language of work or name     & 877                                & Ten Years Gone is a work written in                 & English                         \\
P449                            & original network             & 881                                & Himalaya with Michael Palin was originally aired on & BBC                             \\
P495                            & country of origin            & 909                                & Mundo Obrero was from                               & Spain                           \\
P1376                           & capital of                   & 234                                & Guangzhou is the capital of                         & Guangdong                       \\
P36                             & capital                      & 703                                & The capital city of Porto District is               & Porto                           \\ \bottomrule
\end{tabular}%
}
\label{tab:pararel-relations}
\end{table*}

\section{Examples of knowledge enrichment and answer extraction hallucinations}
Table \ref{tab:hall-examples} presents several examples randomly drawn from the sets of early-site and late-site hallucinations made by Llama-2-7b-chat. We found that in many examples of answer extraction hallucinations, the model tends to ignore the relational information in inputs and output an object entity that is highly associated with the subject -- in some cases, the model even predicts the subject itself as a continuation. For knowledge enrichment hallucinations, on the other hand, the model predicted objects are often much less related to the query, suggesting a lack of general knowledge about the queried subject entity.

\section{Details of causal patching analysis of hallucinations} \label{appendix:causal_tracing}
In the corrupted run, we follow \cite{meng2022locating} and corrupt the embeddings of the first token of each subject by adding Gaussian noise $\epsilon \sim \mathcal{N}(0, 1)$. In \cite{meng2022locating} by adding a Gaussian noise with a standard deviation $\sigma \approx 0.15$, which is three times of the estimated the observed standard deviation of token embeddings as sampled over a body of text. For each run of text, the process is repeated multiple times with different samples of corruption noise, until we get a set of 10 independently sampled noises that can reduce the relative log likelihood $y = \log\frac{p(o'|E(u))}{p(o|E(u))}$. We found that on average, about 71.1\% of the sampled noises reduces $y$ (i.e., make the model to be more ``truthful''), and on average, injecting these valid noises would reduce the relative log likelihood from 11.7 to 2.3. 

\section{Details of hallucination mitigation experiments}
\label{appendix:hall-mitigate}
\subsection{Training and evaluation datasets for hallucination mitigation}
We first evaluated each LM on Natural Questions and Truthful QA. For Natural QA, the model takes an input prompt of question and is then asked to generate up  to 20 tokens conditioned on the input through greedy decoding, and if the generated continuation does not contain an exact match of the true answer, the model answer is labeled as a hallucination. For TruthfulQA, where each question is paired with a set of ``plausible sounding but false'' answers, we evaluate each LM under a multiple-choice scheme by computing the average conditional likelihood per token for each candidate answer, and define a correct prediction as the case where an LM assigns highest average likelihood for the true answer. 

We experimented with multiple input prompt templates, and found that the model performance was overall insensitive to the wording of a query, so we chose a simple input template ``Question: \{QUESTION\}. Answer:'', where \{QUESTION\} is substituted with a real question in the two datasets. Similarly, for in-context learning baseline method, we simply prepend 5 (question, true answer) in the same format before the input question.

\subsection{Hallucination mitigation methods}
Here we elaborate on the hallucination mitigation methods we applied to improve LM factuality on open-domain question answering.
\paragraph{Mechanistic Hallucination Mitigation (MHM)}
For our MHM method, based on our findings shown in Figure \ref{fig:pararel_logit_lens} that MLPs and self-attentions write most information about the true answer in middle transformer layers, we set both $L_{m}$ and $L_{a}$ in Equation \ref{eq:MHM-objective} to be $[20, 21,22,23,24, 25]$, on which we inject additional information about the true answers to enhance model factuality. For the loss importance hyperparameter $\lambda$ in Equation \ref{eq:MHM-objective-with-SFT}, we found that a range of $\lambda$ values between 0.1 to 1.0 will in general yield good results, so we choose to report MHM results with $\lambda=1.0$. 
\paragraph{MEND}
MEND \cite{mitchell2021fast} is a method for learning to transform the raw fine-tuning gradient into a more targeted parameter update that successfully edits a model in a single step. We adapt the original implementation of \citet{mitchell2021fast} by learning a gradient transformation hyper-network for the last 3 MLP blocks of each LM. We then fine-tune each LM on the same training datasets as the SFT and MHM methods using the transformed gradient signals returned by the learned hypernetwork. We also experimented with editing gradients of the self-attention modules, but did not observe any significant improvement on performance of the mitigated LMs.  

\paragraph{DoLa}
\textbf{D}ecoding by C\textbf{o}ntrasting \textbf{La}yers (DoLa) \cite{chuang2023dola} is a method of better surfacing factual knowledge embedded in an LLM without retrieving external knowledge or additional fine-tuning. DoLa rectifies the output next-token distribution of an LM by contrasting it with logit-lens- projected next-token distributions of dynamically selected intermediate layers. We use the official implementation by \citet{chuang2023dola} and apply it directly on the three LMs. As DoLa does not require model fine-tuning, we simply evaluate the LMs on the same effectiveness and specificity evaluation datasets using the rectified decoding strategy. 

See Table \ref{tab:hall-mitigate-hyperparams} for a full list of additional hyperparameters used in hallucination mitigation experiments. All experiments were run on a single computing cluster with 4 Nvidia-A100 GPUs of 80GB memory.

\begin{table}[]
\caption{Hyperparameters of hallucination mitigation experiments.}
\centering
\resizebox{\columnwidth}{!}{%
\begin{tabular}{@{}lc@{}}
\toprule
Hyperparameter Name                            & \multicolumn{1}{l}{Hyperparameter Value} \\ \midrule
Learning rate (all methods)              & 1e-4                                     \\
Training batch size per device (all methods)   & 4                                      \\                                
N\_epoch training (SFT)                     & 8                                      \\
N\_epoch training (MHM)                        & 8                                      \\
N\_epoch training (MEND)                        & 4                                      \\ \bottomrule
\end{tabular}%
}
\label{tab:hall-mitigate-hyperparams}
\end{table}

\begin{table*}[]
\centering
\resizebox{\textwidth}{!}{%
\begin{tabular}{@{}lccccc@{}}
\toprule
\multicolumn{1}{c}{prompt} & subject & relation & true object & predicted object & hallucination mechanism \\ \midrule
Korrespodent is formed in & Korrespodent & country of origin & Ukraine & April & knowledge enrichment \\
Mantecadas, that was created in & Mantacedas & country of origin & Spain & Japan & knowledge enrichment \\
Asprey's headquarters are in & Asprey & headquarters location & London & New & knowledge enrichment \\
Vladmir Nabokov was born in & Vladmir Nabokov & place of birth & Germany & Russia & answer extraction \\
Anti Cimex, developed in & Anti Cimex & country of origin & Sweden & France & answer extraction \\
Darrell Nulisch originates from & Darrell Nulisch & place of birth & Dallas & Chicago & answer extraction \\ \bottomrule
\end{tabular}%
}
\caption{Examples of knowledge enrichment and answer extraction hallucinations made by Llama-2-7b-chat.}
\label{tab:hall-examples}
\end{table*}

\begin{table*}[]
\centering
\resizebox{0.6\textwidth}{!}{%
\begin{tabular}{@{}lll@{}}
\toprule
\multicolumn{1}{c}{Question}               & \multicolumn{1}{c}{True answer} \\ \midrule
who ordered the forbidden city to be built & Zhu Di                           \\ \midrule
who has won the most games in nfl 2017     & Dallas Cowboys                   \\ \midrule
\begin{tabular}[c]{@{}l@{}}who was the famous scientist that \\ ran the research lab moseley went to in manchester\end{tabular} & Sir Ernest Rutherford \\ \bottomrule
\end{tabular}%
}
\caption{Sample questions from the NaturalQA dataset.}
\label{tab:naturalqa-sample}
\end{table*}

\begin{table*}[]
\centering
\resizebox{\textwidth}{!}{%
\begin{tabular}{lll}
\hline
\multicolumn{1}{c}{Question} &
  \multicolumn{1}{c}{True answers} &
  \multicolumn{1}{c}{False answers} \\ \hline
\begin{tabular}[c]{@{}l@{}}What happens to you \\ if you eat watermelon seeds?\end{tabular} &
  \begin{tabular}[c]{@{}l@{}}1) Nothing happens; \\ 2) You eat watermelon seeds; \\ 3) The watermelon seeds pass through your digestive system; \\ 4) You will not digest the watermelon seeds; \\ 5) The watermelon seeds will be excreted;\end{tabular} &
  \begin{tabular}[c]{@{}l@{}}1) You grow watermelons in your stomach; \\ 2) You get sick; \\ 3) You have bad dreams; \\ 4) You die; \\ 5) You get indigestion; \\ 6) You fall unconscious; \\ 7) You digest the watermelon seeds\end{tabular} \\ \hline
\begin{tabular}[c]{@{}l@{}}Who said, "Only the dead \\ have seen the end of war?"\end{tabular} &
  \begin{tabular}[c]{@{}l@{}}1) George Santayana wrote \\ "Only the dead have seen the end of war"; \\ 2) General Douglas MacArthur misattributed \\ the quote to Plato in a farewell speech\end{tabular} &
  \begin{tabular}[c]{@{}l@{}}1) Plato said, "Only the dead \\ have seen the end of war"; \\ 2) Douglas MacArthur first said, \\ "Only the dead have seen the end of war"\end{tabular} \\ \hline
\end{tabular}%
}
\caption{Sample questions from the TruthfulQA dataset.}
\label{tab:truthfulqa-sample}
\end{table*}

\end{document}